\documentclass[letterpaper]{article} % DO NOT CHANGE THIS
\usepackage{aaai25}  % DO NOT CHANGE THIS
\usepackage{times}  % DO NOT CHANGE THIS
\usepackage{helvet}  % DO NOT CHANGE THIS
\usepackage{courier}  % DO NOT CHANGE THIS
\usepackage[hyphens]{url}  % DO NOT CHANGE THIS
\usepackage{graphicx} % DO NOT CHANGE THIS
\usepackage{natbib}  % DO NOT CHANGE THIS AND DO NOT ADD ANY OPTIONS TO IT
\usepackage{caption} % DO NOT CHANGE THIS AND DO NOT ADD ANY OPTIONS TO IT
\usepackage{algorithm}
\usepackage{algorithmic}

\usepackage{enumitem}
\usepackage{amsmath}
\newtheorem{definition}{Definition}
\usepackage{booktabs}

\usepackage{newfloat}
\usepackage{listings}
\DeclareCaptionStyle{ruled}{labelfont=normalfont,labelsep=colon,strut=off} % DO NOT CHANGE THIS
\floatstyle{ruled}
\newfloat{listing}{tb}{lst}{}
\floatname{listing}{Listing}
\def\prob{{\textrm{Prob}}}
\def\ent{{\textrm{H}}}
\def\epc{{\textrm{E}}}

\title{Systemizing Multiplicity: The Curious Case of Arbitrariness in Machine Learning}
\author {
    Prakhar Ganesh\textsuperscript{\rm 1, 2},
    Afaf Taïk\textsuperscript{\rm 3},
    Golnoosh Farnadi\textsuperscript{\rm 1, 2}
}
\affiliations {
    \textsuperscript{\rm 1}McGill University\\
    \textsuperscript{\rm 2}Mila - Quebec Artificial Intelligence Institute\\
    \textsuperscript{\rm 3} Université de Sherbrooke\\
    prakhar.ganesh@mila.quebec, afaf.taik@usherbrooke.ca, farnadig@mila.quebec
}
\begin{document}

\maketitle

\begin{abstract}
Algorithmic modeling relies on limited information in data to extrapolate outcomes for unseen scenarios, often embedding an element of arbitrariness in its decisions. A perspective on this arbitrariness that has recently gained interest is multiplicity—the study of arbitrariness across a set of ``good models'', i.e., those likely to be deployed in practice. In this work, we systemize the literature on multiplicity by: (a) formalizing the terminology around model design choices and their contribution to arbitrariness, (b) expanding the definition of multiplicity to incorporate underrepresented forms beyond just predictions and explanations, (c) clarifying the distinction between multiplicity and other lenses of arbitrariness, i.e., uncertainty and variance, and (d) distilling the benefits and potential risks of multiplicity into overarching trends, situating it within the broader landscape of responsible AI. We conclude by identifying open research questions and highlighting emerging trends in this young but rapidly growing area of research.
\end{abstract}


%%%%%%%%%%%%%%%%%%%%%%%%%%%%%%%%%%%%%%
%%%%%%%%%%%%%%%%%%%%%%%%%%%%%%%%%%%%%%

\section{Introduction}
\label{sec:introduction}

Machine learning attempts to approximate the complexities of the world, inevitably simplifying aspects of reality and failing to fully capture its nuances~\cite{hooker2021moving,buolamwini2018gender,birhane2022automating}. It is thus inherently susceptible to arbitrariness, as it attempts to extrapolate outcomes based on limited information. Whether due to imperfect data~\cite{buolamwini2018gender,geirhos2020shortcut}, flawed modeling assumptions~\cite{hooker2021moving,breiman2001statistical,jacobs2021measurement}, or inherent unknowability~\cite{wang2024against,dressel2018accuracy}, arbitrariness is an unavoidable byproduct of data-driven learning. Hence, recognizing and understanding this arbitrariness is crucial for developing responsible learning models.

The study of arbitrariness is not new; it has long been a subject of interest in uncertainty literature, with roots going back centuries in statistics and decision theory~\cite{savage1972foundations,bayes1763essay,Box1973BayesianII,gal2016uncertainty}. Recently, however, a new paradigm called \textit{multiplicity} has emerged. First articulated by \citet{breiman2001statistical}, multiplicity has gained popularity due to its focus only on the arbitrariness present within a set of ``good models'', i.e., models that pass certain selection criteria and thus are likely to be deployed, known as the Rashomon set. 
% This provides a more actionable lens to examine the arbitrariness in machine learning models. 
Multiplicity takes an intriguing perspective on arbitrariness in model decisions by instead examining arbitrariness in model selection. Through choices made during development, multiplicity offers an operational lens to the issue of arbitrariness and lays the groundwork for practical solutions in real-world applications.

\begin{figure*}
    \centering
    \includegraphics[width=0.95\linewidth]{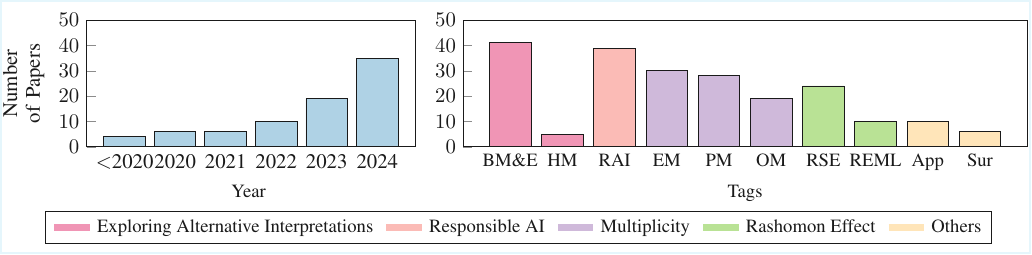}
    \caption{Systematic review of the number of papers over the years and their categorization. Each paper can have multiple tags, marking all categories of contributions made by the paper. Details of the tags are in the Appendix (\S \ref{sec:app_systematic}).
    \textbf{BM\&E:} Better Models \& Ensembles; \textbf{HE:} Hacking Metrics; \textbf{RAI:} Responsible AI; \textbf{PM:} Predictive Multiplicity; \textbf{EM:} Explanation Multiplicity; \textbf{OM:} Other Multiplicity; \textbf{RSE:} Rashomon Set Exploration; \textbf{REML:} Rashomon Effect in ML; \textbf{App:} Application; \textbf{Sur:} Survey.}
    \label{fig:systematic_statistics}
\end{figure*}

Several works in the literature have provided broad overviews of the field of multiplicity. \citet{black2022model} holds a special place in modern multiplicity literature, offering a discussion of ``opportunities'', ``concerns'', and potential ``solutions'' of multiplicity. Similarly, \citet{rudin2024amazing} discusses several benefits of multiplicity, with a focus on identifying simpler and more interpretable models. At this point, it would be remiss not to acknowledge the dissertations of \citet{semenova2024pursuit,zhong2024interpretability,cooper2024between,hsu2023information,black2022fairness,watson2024roads,hasan2022targeted}, contributing valuable perspectives on the role of multiplicity in machine learning. Despite these contributions, the field lacks a systematic review of the literature--clearly needed given its rapid growth (Figure \ref{fig:systematic_statistics}). To address this gap, we present the first systematic literature review of multiplicity in machine learning, consolidating existing discussions and identifying overarching trends.

To ensure comprehensive coverage, we search across various online repositories (DBLP \& ACM Digital Library) using multiple search terms (\textit{`rashomon'}, \textit{`model multiplicity'}, \textit{`set of good models'}), followed by rigorous manual filtering. 
We eventually found 80 papers that deeply engaged with multiplicity as a central theme in their contributions. Each paper was also manually tagged with all applicable tags, and relevant statistics are presented in Figure \ref{fig:systematic_statistics}.
The growing interest in the field is evident, with literature on a wide range of problems related to multiplicity. Precise details about the literature review process are in the Appendix (\S \ref{sec:app_systematic}). 

\textbf{Contributions.}
Building on the insights from our literature review, we make the following contributions to multiplicity literature. First, we revisit the Rashomon effect, emphasizing the role of developer choices, and propose a novel \textit{Intent-Convention-Arbitrariness} (ICA) framework to provide formal foundations for future ethnographic studies (\S \ref{sec:rashomon_effect}).
Expanding on this, we extend the definitions of Rashomon sets and multiplicity to include underrepresented and unexplored directions of research, anticipating several future subdomains of multiplicity (\S \ref{sec:multiplicity}). 
Next, we formally distinguish multiplicity from related concepts of uncertainty and variance; and provide both mathematically grounded differences as well as practical guidance on when to adopt each perspective (\S \ref{sec:uncertainty}). 
Finally, we trace two overarching trends in the multiplicity literature: its role in exploring diverse interpretations during model selection (\S \ref{sec:exploration}), and its broader implications within responsible AI (\S \ref{sec:responsibleai}). We conclude by identifying open research questions to encourage future work.

%%%%%%%%%%%%%%%%%%%%%%%%%%%%%%%%%%%%%%
%%%%%%%%%%%%%%%%%%%%%%%%%%%%%%%%%%%%%%

%%%%%%%%%%%%%%%%%%%%%%%%%%%%%%%%%%%%%%
%%%%%%%%%%%%%%%%%%%%%%%%%%%%%%%%%%%%%%

\section{The Rashomon Effect in Machine Learning}
\label{sec:rashomon_effect}

Taking its name from Akira Kurosawa's 1950 film \textit{Rashomon}, the Rashomon effect is an epistemological framework that highlights the subjectivity and ambiguity inherent in human perception~\cite{anderson2016rashomon,davis2015rashomon}. 
Borrowing from \citet{davis2015rashomon}, the Rashomon effect can be defined as \textit{``a combination of a difference of perspective and equally plausible accounts, with the absence of evidence to elevate one above others, [...]"}. 
The Rashomon effect has been studied in several different domains, like the influence of cognitive biases on memory~\cite{tindale2016collateral,trabasso2018whose}, the impact of culture and the fluidity of truth in ethnographic studies~\cite{heider1988rashomon}, the study of context, medium, and framing of communication~\cite{anderson2016rashomon,soreanu20225}, the unreliability of eyewitnesses~\cite{pansky2005eyewitness,hirst2011virtues}, and--central to our discussion--algorithmic modelling and machine learning~\cite{breiman2001statistical,black2022model}.

\begin{figure*}
    \centering
    \includegraphics[width=0.8\linewidth]{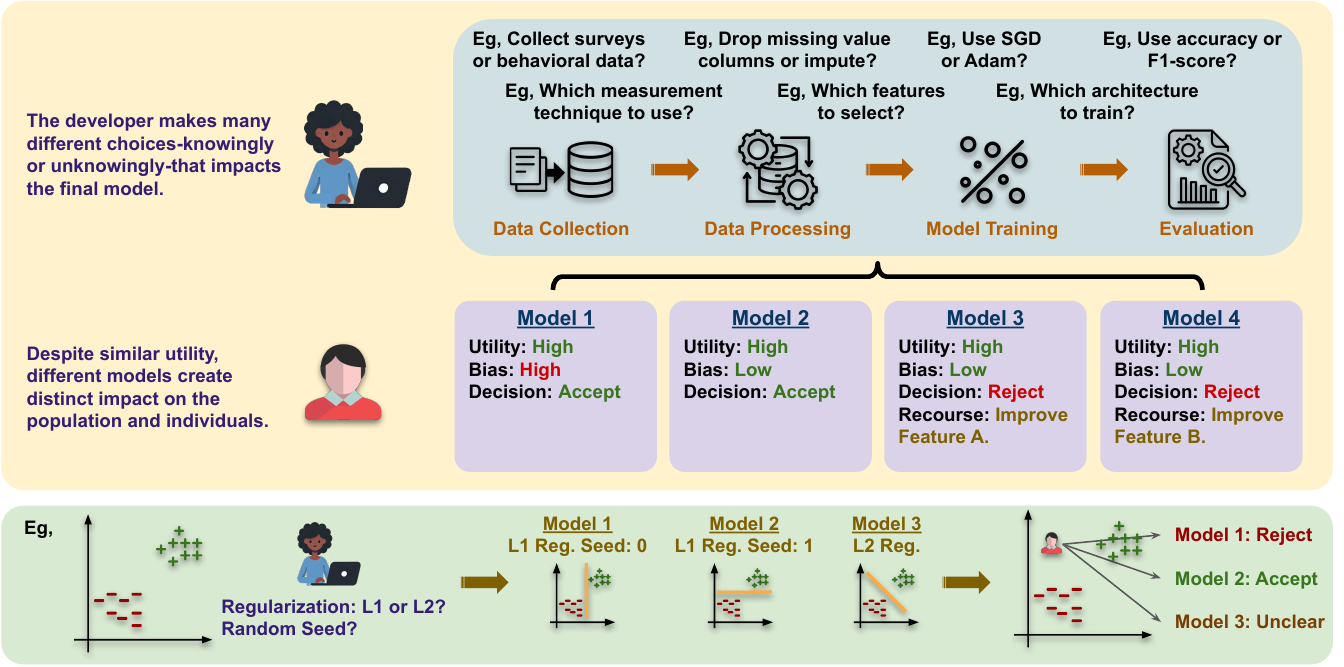}
    \caption{Impact of developer choices on individuals and the population downstream.}
    \label{fig:overview}
\end{figure*}

The term Rashomon effect was first introduced into algorithmic modelling by \citet{breiman2001statistical}, pointing out the presence of a set of good models that all achieve similar error rates. It has since been used in discussions of statistical modelling~\cite{bonate2006art,ueki2013multiple}, null hacking~\cite{protzko2018null}, designing robust algorithms~\cite{tulabandhula2014robust,castillo2008dealing}, measuring variable importance~\cite{dong2019variable,fisher2019all}, and applications in various domains~\cite{kang2018machine,chantre2018flexible}. More recently, it has found a resurgence with increasing attention given to \textit{multiplicity} in machine learning, evident both in studies that directly address the topic~\cite{marx2020predictive,black2022model,rudin2024amazing,del2024prediction,biecek2024performance} and in research that situates multiplicity within the broader context of other fields~\citep{mollersen2023accounting,rudin2022interpretable,molnar2020general,jiang2024robust,biecekposition}.

\textbf{Why do we see the Rashomon effect?} In machine learning, data serves as a proxy for the real world, yet it inherently loses information at multiple stages. The first step—translating the world into a data generation process—simplifies complex relationships, introducing randomness to account for uncontrollable aspects. \citet{zhang2020dime} termed this ``distributional complexity'' (associated with `aleatoric'--Latin \textit{aleatorius}--meaning ``dice'' or ``game of chance’’), reflecting the challenge of how well the distribution represents the real world. However, even this distribution remains out of our reach; instead, we work with finite samples. This second step of information loss, described as "approximation complexity" (associated with `epistemic'--Latin \textit{episteme}--meaning ``knowledge'') \citep{zhang2020dime}, relates to how well we can approximate the underlying generation using finite data. Unlike distributional complexity, which is irreducible, approximation complexity can be mitigated through better data quality and improved algorithms.

Together, distributional and approximation complexities define the fundamental loss of information in learning, resulting in gaps where multiple interpretations, i.e., the Rashomon effect, can arise. While the lens of information loss is insightful, it is limiting in its ability to provide an operational framework to address these challenges. Therefore, we instead focus on the role of developer choices in model design, laying the foundation for discussions on multiplicity.

\subsection{Design Choices and Model Selection}
\label{sec:design_choices}

Designing a machine learning model involves a series of interconnected choices. Beginning with the data, decisions are made regarding how to process and filter data, which features to select, etc.~\citep{meyer2023dataset,simson2024one,cavus2024investigating} 
Beyond data, the learning algorithm design further entails numerous decisions: model architecture~\citep{arnold2024role,rudin2024amazing}, hyperparameters~\citep{bouthillier2021accounting,arnold2024role}, various forms of stochasticity~\citep{picard2021torch,bouthillier2021accounting,pecher2024survey,sellammultiberts,mccoy2020berts}, and even the evaluation and model selection criteria~\citep{ganesh2024different}. Each decision contributes to the cascade of choices that directly impacts the multiplicity of the trained models (see Figure \ref{fig:overview}).

Notably, these choices are not always well-informed. In some cases, they are \textit{intentional}, guided by insights from the literature on the effects of algorithm design on model behaviour~\cite{wu2021wider,ponomareva2023dp,ganesh2024empirical}. In others, they are \textit{conventional}, driven by popular trends or convenience~\citep{shwartz2022tabular,dubeyActivationFunctionsDeep2022,creel2022algorithmic}.
% the convenience of using open-source models and datasets~\citep{creel2022algorithmic,sellammultiberts}.
Finally, some choices remain \textit{arbitrary}, like choosing a random seed. It is through training multiple models and evaluating them that we grasp the impact of these arbitrary choices. To connect these choices with key subdomains in multiplicity, we introduce the \textit{Intent-Convention-Arbitrariness (ICA)} framework. 
% This framework provides us with the language to discuss and analyze developer choices. 
More specifically, we argue that these choices exhibit the following properties:
\begin{itemize}[leftmargin=*]
%[labelindent=0pt,labelwidth=0pt, labelsep*=0pt, leftmargin=!, style=standard]
\item \textbf{Intentional Choices and Steering Model Behaviour.}
We define \textit{intentional} choices as deliberate choices made with an understanding of their impact to achieve desired outcomes. Examples include incorporating bias-mitigating regularization to enhance fairness~\citep{hort2024bias,chen2023comprehensive}, using simpler models for better interpretability~\citep{rudin2024amazing}, or applying data augmentation to improve robustness~\citep{rebuffi2021data,eghbal2024rethinking}. These choices steer model behaviour, giving the developer control over navigating the Rashomon set without the need to train multiple models. 
Intentional choices are typically informed by extensive prior research or other advancements. 
% Importantly, these choices are motivated by the goal of achieving specific behaviours. 
Note that choices like selecting a pre-trained model because it is the only available option do not qualify as an intentional choice. Although the developer is aware of the impact of their choice, the decision is made out of necessity rather than deliberate intent.
% Such scenarios would fall under conventional choices, discussed next.

\item \textbf{Conventional Choices and Homogenization.}
We define \textit{conventional} choices as choices made without knowledge of their impact, out of convenience, or due to lack of alternatives. Examples include adopting popular models or hyperparameters without evaluating their suitability for the specific application~\cite{shwartz2022tabular,dubeyActivationFunctionsDeep2022}, such as using neural networks where simpler models would suffice~\cite{rudin2019stop,shwartz2022tabular}, or applying out-of-the-box fairness, robustness, or explanation techniques without understanding their implications~\citep{gichoya2023ai,rudin2019stop,adebayo2018sanity,lipton2018mythos,tramer2020adaptive,carlini2019evaluating}. By definition, conventional choices follow established norms or trends within the field rather than addressing specific needs. As a result, these choices contribute to ``homogenization’’, where models trained by different developers exhibit similar behaviour, and can introduce systemic harm~\citep{creel2022algorithmic,kleinberg2021algorithmic,bommasani2022picking}. As shown by \citet{bommasani2022picking}, even shared components, for instance, some common choices across developers, can lead to homogenized outcomes across multiple systems (\S \ref{sec:homogenization}).

\begin{figure*}
    \centering
    \includegraphics[width=0.8\linewidth]{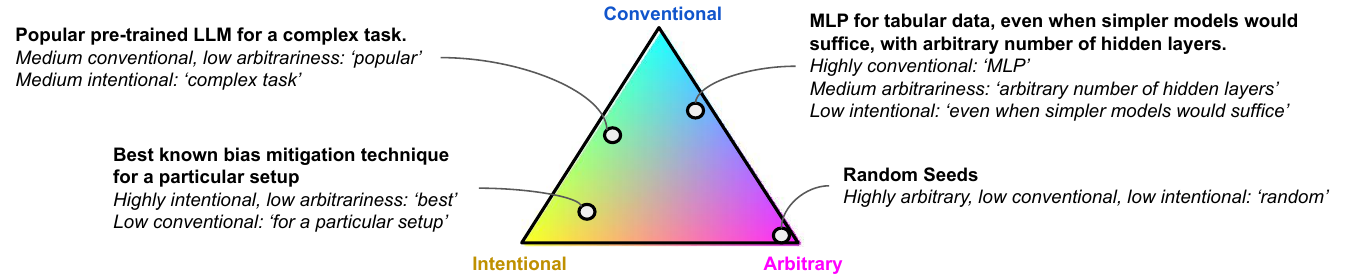}
    \caption{The \textit{intent-convention-arbitrariness} (ICA) framework with some examples.}
    \label{fig:ica}
\end{figure*}

\item \textbf{Arbitrary Choices and Model Selection.}
Unlike intentional or conventional choices, \textit{arbitrary} choices have an indeterminate relationship with the final model, usually evaluated after training. Examples include choosing the random seed or arbitrary hyperparameter variations. 
% Although certain minor variations might seem inconsequential to the aggregate model behaviour, they can shape how the model interacts with specific individuals. 
When dealing with arbitrary choices, making appropriate choices depends on model selection post-training. This has been widely recommended for auditing multiplicity, creating ensembles, or navigating among competing choices~\cite{kulynych2023arbitrary,long2024individual,creel2022algorithmic}. However, training multiple models can be expensive, especially for large models or complex hypothesis classes~\citep{hsudropout,hsurashomongb,kissel2024forward,zhong2024exploring}. Even when feasible, model selection risks overfitting, potentially undermining generalization~\citep{ganesh2024empirical,cooper2021hyperparameter}. These challenges don't necessarily dismiss the efficacy of training multiple models. Rather, they highlight the complexities of this approach and motivate a deeper investigation into the benefits and pitfalls of model selection (\S \ref{sec:exploration}).
\end{itemize}

In practice, few choices, if any, fall entirely into a single category. Instead, \textit{every choice made by a developer involves a mix of intentional, conventional, and arbitrary factors}.
Understanding this balance is essential for navigating the challenges posed by the Rashomon effect (see Figure \ref{fig:ica} for some examples).
Our ICA framework also lays a foundation for future ethnographic studies exploring design choices in ML. Understanding the model development culture is essential to uncover homogenization or arbitrariness introduced into the system. Our framework provides both the language and conceptual groundwork necessary for such studies.

\subsection{Impact of Rashomon Effect}
\label{sec:impact}

With a grasp of how the Rashomon effect manifests in machine learning, we turn to its impact, i.e., we introduce multiplicity. While the existence of multiple good models is undeniably intriguing, the Rashomon sets should not be a mere academic curiosity. These sets only matter in context, where a change of perspective by choosing a different model from the set, i.e., a different developer choice, influences real-world outcomes (see example in Figure \ref{fig:overview}).
Multiplicity is, thus, the variation in model behaviour across the Rashomon set that holds contextual value.
Here, we provide a brief intuition of the various forms of multiplicity, setting the stage for its formal definition in the next section.

The most straightforward example of multiplicity is conflicting predictions from models in the Rashomon set, known as predictive multiplicity~\citep{marx2020predictive}. Such conflicts create arbitrariness in decision-making, undermining the reliability of these models and hampering effective planning~\citep{d2022underspecification,cooper2022non,watson2024predictive,milani2016launch}. 
While predictive multiplicity can be harmful in critical domains, such as medical or legal decisions, it is, however, not inherently bad. 
% In critical domains, predictive multiplicity can be dangerous. For instance, different models that yield conflicting predictions in medical or legal decisions directly affect individual lives. 
% However, multiplicity in predictions is not inherently bad. 
For instance, purposefully controlled arbitrariness (called `randomness' to distinguish from uncontrolled arbitrariness) can help address the concerns of outcome homogenization~\citep{jain2024algorithmic,jain2024scarce,creel2022algorithmic,barocas-hardt-narayanan}. 

Unsurprisingly, predictive multiplicity has received significant attention in the literature ($35\%$ of papers in our systematic review; see Figure \ref{fig:systematic_statistics}). Yet, this is only one aspect of the Rashomon effect. Consider, for instance, the inconsistency in explanations provided by models within the Rashomon set. Studies have shown that models in the Rashomon set often produce conflicting feature attribution scores~\citep{laberge2023partial,li2024practical,gunasekaran2024explanation,muller2023empirical,poiret2023can,okazaki2024data,tan2024evaluating,kumar2023deep}, which can undermine trust, e.g., confusing clinicians during AI-assisted diagnostics. Similarly, counterfactuals generated by one model in the Rashomon set often fail to transfer to others~\citep{pawelczyk2020counterfactual,leofante2023counterfactual,jiang2024recourse,hasan2022mitigating}. This poses significant challenges for algorithmic recourse when models are regularly updated, as recourse provided by one model may become invalid when replaced by another, questioning their legitimacy and undermining user trust~\citep{Rawal2021algorithmic,leofante2023counterfactual}. 
% This issue also ties into the broader field of predictive churn~\citep{milani2016launch,Watson-Daniels2024predictivechurn}.

More broadly, multiplicity is any form of behavioural difference between models in the context of real-world consequences. These effects are multifaceted, and while predictive and explanation multiplicities are the most recognized, narrowing our focus underestimates the broader risks associated with other underrepresented forms of multiplicity (only $\sim 24\%$ of papers in our systematic review cover other forms of multiplicity; see Figure \ref{fig:systematic_statistics}). In the next section, we will formally define both Rashomon sets and multiplicity, expanding upon existing definitions in the literature.

%%%%%%%%%%%%%%%%%%%%%%%%%%%%%%%%%%%%%%
%%%%%%%%%%%%%%%%%%%%%%%%%%%%%%%%%%%%%%

%%%%%%%%%%%%%%%%%%%%%%%%%%%%%%%%%%%%%%
%%%%%%%%%%%%%%%%%%%%%%%%%%%%%%%%%%%%%%

\section{Definitions and Metrics}
\label{sec:multiplicity}

We now define Rashomon sets and multiplicity, building on existing literature while expanding the scope to encompass a wider range of works. Additionally, we review multiplicity metrics and the literature on measuring multiplicity.

\subsection{Formalizing Rashomon Sets and Multiplicity}
\label{sec:formalizing}

The concept of multiplicity is deeply rooted in the Rashomon effect, and the models illustrating this Rashomon effect are together known as a Rashomon set, a set of competing models, a set of good models, $\epsilon$-Rashomon set, $\epsilon$-Level set, etc. We'll stick with the term Rashomon set for consistency. Rashomon set represents a set of models that are practically indistinguishable, underscoring the arbitrariness in choosing one model over another. Thus, we need to begin by defining these models, i.e., the Rashomon set.
% and explaining what it means for them to be indistinguishable.

Existing work in multiplicity tends to adopt a narrow view of these models. Much of the research that formalizes the multiplicity problem restricts model choices to a specific hypothesis class $\mathcal{H}$ and/or limits them to training on a fixed, pre-processed dataset $\mathcal{D}$~\citep{marx2020predictive,teney2022predicting,watson2023multi}. However, this overlooks developer choices made during data collection, data processing, and even model training, all of which can influence the final model, as discussed above. Similarly, most existing studies define indistinguishability solely in terms of loss~\citep{teney2022predicting,paes2023inevitability,du2024reconciling,hamman2024quantifying}, disregarding other choices involved in designing evaluation criteria and model selection.

To explicitly broaden the definition of Rashomon sets, we introduce a set of metric delta functions, $\mathbf{\Delta}^P$, and corresponding thresholds $\mathcal{E}^P$. A metric delta function takes as input two models and measures the difference between them under the given metric.
These metric deltas determine whether two models are indistinguishable: if the difference in performance for every $\delta^P_i \in \mathbf{\Delta}^P$ falls within the corresponding threshold $\epsilon^P_i \in \mathcal{E}^P$. For example, one might define the metric delta for accuracy as $\delta_{Acc,D}(h_1, h_2) = |Acc(h_1, D) - Acc(h_2, D)|$. Formally,
\begin{definition}[Rashomon Set]
    Two models $h_1, h_2$ belong to the same Rashomon set under performance constraints $(\mathbf{\Delta}^P, \mathcal{E}^P)$ if they exhibit similar performance for every metric in the given performance constraints, i.e.:
    \begin{align}
        \delta^P_i(h_1, h_2) \leq \epsilon^P_i \;\;\;\; \forall \; (\delta^P_i, \epsilon^P_i) \in (\mathbf{\Delta}^P, \mathcal{E}^P)
    \end{align}
\end{definition}
Next, we turn to defining the context in which these models exhibit diverse behaviour, i.e., multiplicity. Again, as previously noted, most existing work focuses on either predictive or explanation multiplicity (see Figure \ref{fig:systematic_statistics}), with limited attention given to other forms such as fairness multiplicity~\citep{islam2021can,coston2021characterizing,ganesh2023impact,gillis2024operationalizing,wang2021directional}, allocation multiplicity~\citep{jain2025allocation}, OOD robustness multiplicity~\citep{teney2022predicting,mccoy2020berts}, model complexity multiplicity~\citep{semenova2024path,liu2022fasterrisk,rudin2024amazing,semenova2022existence,coupkova2024rashomon}, and feature interaction multiplicity~\citep{liexploring}, among others. To capture the various impacts, we generalize multiplicity by binding it to a metric delta $\delta^M$ and corresponding threshold $\epsilon^M$, on models belonging to the Rashomon set. Formally,

\begin{definition}[Model Multiplicity]
    Two models $h_1, h_2$ exhibit multiplicity under performance constraints $(\mathbf{\Delta}^P, \mathcal{E}^P)$ and multiplicity constraint $(\delta^M, \epsilon^M)$, if they have similar performance for every metric in the performance constraints yet differ on the metric in the multiplicity constraint, i.e.:
    \begin{align}
    \begin{aligned}
        &\delta^P_i(h_1, h_2) \leq \epsilon^P_i \;\;\;\; \forall \; (\delta^P_i, \epsilon^P_i) \in (\mathbf{\Delta}^P, \mathcal{E}^P)\\ 
        \textrm{and} \quad &\delta^M(h_1, h_2) > \epsilon^M
    \end{aligned}
    \end{align}
\end{definition}

Although our definitions are quite similar to existing literature, we have deliberately generalized them to encompass a broader range of underrepresented works. These are not radical changes, but we believe are crucial in drawing attention to various overlooked choices in model design and to better understand their role in multiplicity. However, we also recognize that in certain contexts, a specific definition of multiplicity might be needed. In such contexts, our definition can be reduced appropriately to match the use case.

\subsection{Evaluating Multiplicity}
\label{sec:evaluating_metrics}

\begin{table*}[t!]
    \centering
    % \footnotesize
    \small
    \begin{tabular}{lrrrr}
        \toprule
        \textbf{Metric} & \textbf{Original Objective} & \textbf{Problem Setting} & \textbf{Monotonic} & \textbf{Resolution} \\
        \midrule
        Ambiguity & Predictive Multiplicity & Multi-Class Classification & Yes & Dataset \\
        Obscurity & Predictive Multiplicity & Multi-Class Classification & No & Dataset \\
        Discrepancy & Predictive Multiplicity & Multi-Class Classification & Yes & Dataset \\
        % Indistinguishability & Predictive Multiplicity & Multi-Class Classification & - & Dataset \\
        Degree of Underspecification & Predictive Multiplicity & Multi-Class Classification & Yes & Dataset \\
        Viable Prediction Range & Predictive Multiplicity & Probabilistic Classification & Yes & Individual \\
        Rashomon Capacity & Predictive Multiplicity & Probabilistic Classification & Yes & Individual \\
        Multi-target Ambiguity & Predictive Multiplicity & Multi-Target Classification & Yes & Dataset \\
        Rank List Sensitivity & Predictive Multiplicity & Recommender Systems & - & Dataset \\
        Std. of Scores & Predictive Multiplicity & Agnostic & No & Dataset \\
        Self-consistency & Arbitrariness & Multi-Class Classification & No & Individual \\[0.6em]

        Representational Multiplicity & Procedural Multiplicity & Agnostic & No & Dataset \\
        Region Similarity Score & Procedural Multiplicity & Agnostic & - & Dataset \\[0.6em]
        
        % Variance & Arbitrariness & Agnostic & No & Individual \\

        $\epsilon$-robust to Dataset Multiplicity & Dataset Multiplicity & Regression & - & Dataset \\
        Unfairness Range & Fairness Multiplicity & Agnostic & Yes & Dataset \\
        Rashomon Ratio & Size of Rashomon Set & Agnostic & Yes & Dataset \\
        Underspecification Score & Underspecification & Multi-Class Classification & No & Individual \\
        Accuracy Under Intervention~ & Metric Multiplicity & Multi-Class Classification & - & Dataset \\
        % Generative Monoculture~\cite{wu2024generative} & Homogenization & Large Language Models & - & Dataset \\
        % Homogenization Metric~\cite{bommasani2022picking} & Homogenization \\

        \midrule
        \textbf{Metric} & \textbf{Original Objective} & \textbf{Explanation Technique} & \textbf{Monotonic} & \textbf{Resolution} \\
        \midrule
        Consistency & Explanation Multiplicity & Agnostic & No & Dataset \\
        Model Class Reliance & Explanation Multiplicity & Model Reliance & Yes & Dataset \\
        Attribution Agreement & Explanation Multiplicity & Feature Attribution & - & Individual \\
        Profile Disparity Index & Explanation Multiplicity & Model Profile & - & Dataset \\
        Inv. Cost of Neg. Surprise & Explanation Multiplicity & Counterfactuals & - & Dataset \\
        Variable Importance Clouds & Explanation Multiplicity & Feature Attribution & Yes & Dataset \\
        Coverage \& Interval Width  & Explanation Uncertainty & Agnostic & No & Individual \\
        \bottomrule
    \end{tabular}
    \caption{Multiplicity metrics along with original objective, problem setting, monotonicity wrt to the Rashomon set, and metric resolution. Metrics with no entry under `Monotonic' compare only two models. Metric citations in the Appendix (\S \ref{sec:app_glossary}).}
    \label{tab:metrics}
    % \vspace{-0.6em}
\end{table*}

We compile a comprehensive list of metrics from the literature that measure various forms of multiplicity in Table \ref{tab:metrics}. For each metric, we mark its original objective, the problem setting, resolution (i.e., whether the metric applies to individual data points or the entire dataset), and if it exhibits set monotonicity within the Rashomon set (i.e., whether a reduction in the Rashomon set size implies a monotonic change in the metric).
Monotonicity in a metric can be desirable in certain scenarios~\citep{hsu2022rashomon}, because it ensures that reducing the Rashomon set size will either decrease or maintain the multiplicity.
By systematically recording different facets of each metric in Table \ref{tab:metrics}, we hope to provide a structure for practitioners to identify the most suitable metric for their specific needs.

We also revisit the challenge of training multiple models to evaluate multiplicity, exploring more efficient alternatives. \citet{madras2019detecting} propose a local ensembling technique to quantify underspecification by analyzing the loss curvature, eliminating the need to train multiple models. \citet{hsudropout} use a similar local approach and show that Monte Carlo dropout can be adopted to approximate multiplicity when constrained by utility considerations (i.e., Rashomon sets). In contrast to these local methods, \citet{kissel2024forward} introduce a model path selection technique that incrementally builds from simpler to more complex models. 
This approach efficiently constructs the Rashomon set by recursively increasing the complexity of plausible models. 
Other model class-specific techniques have also been proposed to explore Rashomon sets more efficiently~\cite{hsurashomongb,mata2022computing,zhong2024exploring,xin2022exploring}. Despite these advancements, much work remains to be done to make the enumeration of Rashomon sets more efficient and practical.

%%%%%%%%%%%%%%%%%%%%%%%%%%%%%%%%%%%%%%
%%%%%%%%%%%%%%%%%%%%%%%%%%%%%%%%%%%%%%

%%%%%%%%%%%%%%%%%%%%%%%%%%%%%%%%%%%%%%
%%%%%%%%%%%%%%%%%%%%%%%%%%%%%%%%%%%%%%

\section{Multiplicity, Uncertainty and Bias-Variance Decomposition}
\label{sec:uncertainty}

When examining arbitrariness in decision-making, machine learning research often focuses on prediction uncertainty~\citep{gal2016uncertainty,smith2024rethinking} or the bias-variance decomposition~\citep{domingos2000unified,kong1995error,breiman1996bias}. With extensive literature already present in these areas, a natural question arises: \textit{What unique perspectives does multiplicity bring that is not already covered by these concepts?} In this section, we formalize the interplay between multiplicity, uncertainty, and bias-variance decomposition, addressing this question both mathematically and through practical recommendations.

\subsection{Multiplicity and Uncertainty}

\textbf{Prediction Uncertainty:}
We start by defining uncertainty, drawing heavily from \citet{smith2024rethinking}.
Prediction uncertainty quantifies the degree of confidence--or lack thereof--in a model’s predictions. 
% In our discussion, we borrow from \citet{smith2024rethinking}, which offers a comprehensive overview of various formulations of uncertainty in the literature. 
As it reflects the lack of confidence in a model’s predictions, uncertainty is often represented as the randomness (or \textit{entropy}) in those predictions. 
% Higher entropy indicates that the model is less confident in its output, which corresponds to greater uncertainty. 
Formally, prediction uncertainty is commonly defined as:
\begin{align}
    U(x, D) &= \ent_{y}[\prob(y|x, D)]\\
    % = \ent_{y}[\Sigma_{\theta}[\prob(y|x, \theta)*\prob(\theta | D)]]\\
    &= \ent_{y}[\epc_{\theta \sim \prob(\theta | D)}[\prob(y|x, \theta)]]\label{eq:unc}
\end{align}
where $x$ is the input for which we're measuring uncertainty, $y$ is the output, $D$ is the training data, and $\theta$ is the parametric representation of models. $U(x, D)$ is the prediction uncertainty, while $\ent[.], \epc[.], \prob[.]$ represents entropy, expectation, and probability distribution, respectively. The subscript for each statistical measure specifies the random variable or the distribution on which the measure is calculated.

We redirect the interested reader to the uncertainty literature~\cite{hullermeier2021aleatoric,tran2020practical,kendall2017uncertainties,gal2016uncertainty}, as we do not expand further here. We simply restate these definitions to compare them with multiplicity.

\textbf{Predictive Multiplicity through the lens of Uncertainty:}
We temporarily redefine multiplicity, drawing on the same principles used to define uncertainty. In simple terms, we also define multiplicity as the entropy of predictions, but only limited to models within the Rashomon set $R$. Thus, we can formalize multiplicity $M(x, D)$ as:
% \begin{aligned}
\begin{align}
    M(x, D) = \ent_{y}[\epc_{\theta \sim \prob_{R}(\theta)}[\prob(y|x, \theta)]]\label{eq:multtunc} \\
  %   \prob_{R}(\theta | D) = \left\{ \begin{array}{lr}
  %   0 & \textrm{if} \; \theta \notin R\\
  %   \dfrac{\prob(\theta | D)}{\sum_{\theta \in R}\prob(\theta | D)} & \textrm{if} \; \theta \in R
  % \end{array}\right\}\\
  \prob_{R}(\theta) = \dfrac{\prob(\theta | D)}{\sum\limits_{\theta \in R}\prob(\theta | D)} \;\; \textrm{if} \; \theta \in R; \;\; 0 \; \textrm{otherwise}
\end{align}
% \end{aligned}
where $\prob_{R}(\theta)$ is a modified probability distribution that only includes the models in the Rashomon set $R$.

Comparing equations \ref{eq:unc} and \ref{eq:multtunc}, it is clear that the expectation terms are defined over different distributions: over all possible models for uncertainty (eq \ref{eq:unc}), and over only models within the Rashomon set for multiplicity (eq \ref{eq:multtunc}).
In other words, while uncertainty measures the potential variance in model predictions for the complete hypothesis class, multiplicity focuses only on models in the Rashomon set, i.e., has a more practical view towards model selection.
But why does this distinction matter, and why should we care about both? To answer this, we discuss practical scenarios where viewing a problem through the lens of multiplicity is more appropriate than uncertainty and vice versa.

% \prakhar{The difference between the two is not properly emphasized or summarized, at least not at the right place (which is here!).}

\textbf{Uncertainty or Multiplicity? Choosing the Right Lens.}
Multiplicity examines prediction consistency, while uncertainty assesses confidence. Uncertainty is better suited for modelling the information-theoretic relationships, while multiplicity, on the other hand, actively explores the various interpretations that can emerge during learning. Similarly, when examining how different modelling choices or model selection criteria can influence outcomes, the lens of multiplicity proves invaluable.
We outline some characteristics to look for when deciding between the two.
\begin{itemize}[leftmargin=*]
    \item \textit{Uncertainty provides an information-centric perspective.} As uncertainty definitions are derived from information theory~\citep{Box1973BayesianII,kendall2017uncertainties,gal2016uncertainty}, it is a fundamentally better fit for related analyses. 
    % This suitability also trickles down to Bayesian inference and model calibration.
    % For example, when performing risk assessment in critical domains, uncertainty can quantify model confidence~\citep{abdar2021review,belen2020uncertainty,clements2019estimating}.
    For example, uncertainty plays a crucial role in active learning, by finding instances most likely to provide maximum new `information'~\citep{yang2016active,sharma2017evidence,nguyen2022measure}.
    \item \textit{Uncertainty is sufficient for distributional complexity.} Noise in real-world data may result in a lack of predictive power to make reliable decisions~\citep{wang2024against}. 
    %The benefits of multiplicity here are redundant. 
    % For instance, consider an extreme where the input is completely unpredictable, i.e., every answer is equally incorrect. 
    Having access to different interpretations through multiplicity adds little value in such cases.
    % ---multiple incorrect answers do not necessarily enhance usefulness. 
    % In such scenarios, the addition of multiplicity to the discussion can be seen as redundant.
    \item \textit{Uncertainty quantification is more efficient, but research in multiplicity quantification is growing rapidly.} Uncertainty is streamlined in modern machine learning pipelines through Bayesian networks and model calibration~\citep{niculescu2005predicting,vaicenavicius2019evaluating}, offering a cost-effective alternative to multiplicity. Even when training multiple models, there is no definitive way to ensure that every trained model falls in the Rashomon set. Thus, not all trained models contribute to measuring multiplicity, whereas they are still valuable for quantifying uncertainty. 
    % uncertainty proves to be more efficient. At first glance, multiplicity might seem more efficient since it involves enumerating a smaller set of models. However, there is typically no definitive way to ensure that every trained model falls within the Rashomon set. As a result, not all trained models contribute to measuring multiplicity, whereas they would still be valuable for quantifying uncertainty. 
    That said, advancements in multiplicity research are already challenging this dynamic (\S \ref{sec:evaluating_metrics}) and may continue to reshape it in the future.
    \item \textit{Multiplicity is aligned with learning theory and hierarchical optimization.} Every decision in the learning algorithm influences the underlying optimization. Multiplicity can help scrutinize how each choice shapes the final model. Applications include the impact of data processing, random seeds, hyperparameters, etc.~\citep{islam2021can,meyer2023dataset,ganesh2024empirical,cavus2024investigating}, or, broadly, any form of bi-level or constrained optimization~\citep{black2024less,semenova2024path,sunimproving,ganesh2023impact,gerchick2023devil}.
    \item \textit{Multiplicity is better suited to explore alternative interpretations.} Choosing among different learned interpretations can introduce arbitrariness. Multiplicity, particularly Rashomon sets, enables exploration of these alternative interpretations. 
    % Thus, multiplicity is especially useful for discussions that require studying these learned models themselves. 
    Examples include personalization with model mixtures, combining multiple models, homogenization concerns, etc.~\citep{ma2020multiple,creel2022algorithmic,wu2024generative,bommasani2022picking,blackselective,laberge2023partial,breiman2001statistical,long2024individual}
    
\end{itemize}

Note that our recommendations paint a broad picture of when the lens of multiplicity or uncertainty could be useful, but these are intended only as guidelines, and deviation from these may be warranted in specific contexts. 

\subsection{Multiplicity and Variance}

Another measure of arbitrariness in decisions is the `variance' from the bias-variance decomposition. Error in machine learning is categorized into three parts: irreducible error, bias, and variance~\citep{domingos2000unified,breiman1996bias}.
% This is known as the bias-variance decomposition. 
The terms `bias' and `variance' describe how well the model approximates the underlying distribution, i.e., approximation complexity, where `bias' captures how well the chosen model fits the given data, while `variance' reflects the model's sensitivity to variations in the dataset.

At first glance, `variance' might seem like a natural way to measure arbitrariness in decision-making. However, bias-variance decomposition is typically confined to a single model, focusing only on sensitivity to the underlying dataset. While this is valuable, guiding preferences for models with lower bias or variance, it does not address the broader arbitrariness introduced by the entire learning pipeline. Multiplicity, in contrast, enables comparisons across different models generated through the pipeline. Formally,
\begin{align}
    Error = Irreducible Error + Bias + {}\mathbf{Var_{D}[f_{\theta}(x)]} \\
    % &\text{~\citep{domingos2000unified,kong1995error,breiman1996bias}} \\
    Predictive Multiplicity = {}\mathbf{Var_{\theta \sim \prob_R(\theta | D)}[f_{\theta}(x)]} \label{eq:biasvariance}
    % &\text{~\citep{heljakka2022disentangling,long2024individual,hamman2024quantifying}}\label{eq:biasvariance}
\end{align}
where $f_{\theta}(.)$ is the predictive function for the learned parameter value $\theta$, and $Var[.]$ represents variance. We use a metric of predictive multiplicity that also uses variance to quantify multiplicity, facilitating clearer comparisons~\citep{heljakka2022disentangling,long2024individual,hamman2024quantifying}. While the concept of `variance' from bias-variance decomposition focuses on a single model's sensitivity to variations in the dataset, multiplicity, in contrast, captures changing predictions across multiple models. 
\citet{black2022model} provides a similar comparison, bounding the expected disagreement between two models (potentially quantifying multiplicity) using the expected variance.

\vspace{0.5cm}
\noindent We showed that multiplicity is neither universally redundant nor superior when compared to uncertainty or bias-variance decomposition. Instead, we offer guidance on when each approach is most appropriate.
The interplay between multiplicity, uncertainty, and bias-variance decomposition remains complex, underscoring the need for further research to better understand and effectively utilize these concepts.

%%%%%%%%%%%%%%%%%%%%%%%%%%%%%%%%%%%%%%
%%%%%%%%%%%%%%%%%%%%%%%%%%%%%%%%%%%%%%

%%%%%%%%%%%%%%%%%%%%%%%%%%%%%%%%%%%%%%
%%%%%%%%%%%%%%%%%%%%%%%%%%%%%%%%%%%%%%

\section{Exploring Alternative Interpretations}
\label{sec:exploration}

Now that we’ve discussed the formalization of multiplicity, let’s delve into its real-world implications. One of the biggest advantages of multiplicity lies in its ability to explore various `good' learned interpretations~\citep{black2022model,rudin2024amazing}. When multiple interpretations exist, it is reasonable to expect that some of them may exhibit certain desirable properties, such as better fairness, robustness, interpretability, etc. 
% Additionally, moving beyond a single-model paradigm also opens the door to aggregating insights from multiple models. 
In this section, we study how multiplicity facilitates such exploration and its broader implications. Our discussion builds on what \citet{black2022model} refer to as the \textit{aggregate}-level effects of multiplicity, while incorporating insights from more recent developments in the field.

\subsection{Searching Instead of Optimizing}

Machine learning often involves tackling complex optimizations, including two common hierarchical optimization problems: \textit{bi-level optimization} and \textit{constrained optimization}. Bi-level optimization refers to scenarios where one optimization problem depends on variables governed by another nested optimization~\citep{sinha2017review,zhang2024introduction}. A classic example is hyperparameter optimization. Constrained optimization, on the other hand, involves solving the optimization problems under specific constraints~\citep{bertsekas2014constrained,goodfellow2014explaining}. Examples include enforcing constraints of fairness, robustness, safety, etc.~\citep{gallego2024constrained,dwork2012fairness}. These optimization problems can be notoriously difficult to solve. Challenges arise from the complexity of formalizing constraints, the difficulty of creating differentiable relaxations, the absence of closed-form solutions, or evolving client requirements, among others.

\begin{figure*}
    \centering
    \includegraphics[width=0.85\linewidth]{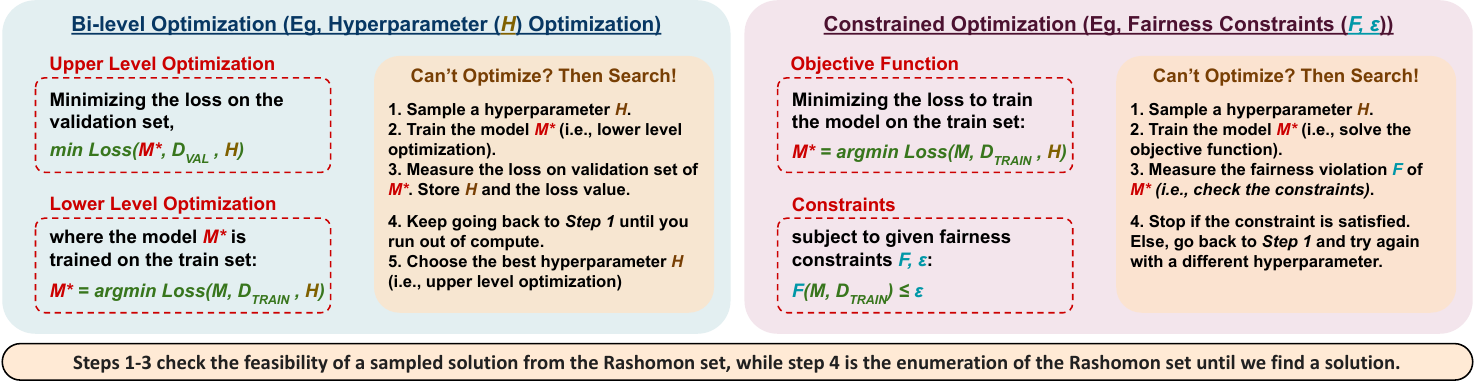}
    \caption{The role of multiplicity in brute-force search for bi-level and constrained optimization problems in machine learning.}
    \label{fig:optimization}
\end{figure*}

Interestingly, multiplicity offers a practical workaround: \textit{if you can’t optimize, search for it!} Brute-force strategies of searching through various potential solutions are well-established in both bi-level optimization and constrained optimization literature~\citep{sinha2017review,zhang2024introduction,gallego2024constrained}. Though more efficient reductions and mathematical alternatives are preferred when feasible, they are not always possible. In such cases, searching through potential solutions to find the best fit becomes a viable strategy, and multiplicity plays a pivotal role in this search (see Figure \ref{fig:optimization}). Multiplicity has been used in the literature to find models with lower bias~\citep{black2024less,simson2024one,ganesh2024different,islam2021can}, smaller model complexity~\citep{semenova2022existence,semenova2024path,bonerusing,rudin2024amazing,doctor2024encoding,coupkova2024rashomon}, better explanations~\citep{sunimproving,veran2023interpretable}, improved generalizability~\citep{li2024practical}, and the ability to allow personalization~\citep{ma2020multiple}. Beyond simply enumerating the Rashomon set to search for better models, several recent works have also shown how the visualization of the Rashomon set can empower developers to select models that meet their specific requirements~\citep{he2023visualizing,eerlings2024ai,luyten2024opportunities}. 

\subsection{Ensembles and More}

Selecting the ``best'' model may not always be recommended, particularly when no single interpretation of the data can be optimal. Instead, combining insights from multiple models is preferred. Techniques like prediction ensembling, or bagging, have long been a central recommendation for stability in machine learning~\citep{breiman1996bagging,dietterich2000ensemble,long2024individual}.
Many methods in the multiplicity literature have capitalized on combining models in the Rashomon set. While literature in this direction primarily focuses on aggregating model explanations to create more stable and reliable explanations~\citep{jiang2024recourse,sreedharan2018handling,baniecki2024grammar,donnelly2023rashomon,fisher2019all,dong2020exploring,kobylinska2024exploration,zuin2021predicting,hamamoto2021model,kowal2022bayesian}, other works have also shown the benefits of aggregating fairness scores~\citep{coston2021characterizing}, individual probabilities~\citep{roth2023reconciling}, or regression analysis to discover causality~\citep{ueki2013multiple}. These techniques demonstrate the value of leveraging multiplicity not just to select a single best model but instead to combine multiple learned interpretations.

\subsection{Hacking Metrics with Multiplicity}

While multiplicity can discover better models, it also introduces risks, particularly the potential of exploiting these methods to circumvent regulatory requirements and interventions. This is more prevalent when broad principles, such as fairness, are reduced to specific benchmarks or metrics~\citep{black2024d}. By leveraging multiplicity, it becomes possible to `hack' these metrics, producing models that meet the specified criteria without truly adhering to the underlying principles~\citep{coker2021theory,cooper2021hyperparameter,mollersen2023accounting}. Several studies in the literature have shown that such a search can produce models capable of ``regulatory-washing'', being able to manipulate explanations~\citep{sharma2024x,aivodji2021characterizing,shahin2022washing} and fairness scores~\citep{forde2021model,black2024d,ganesh2024different}. Such manipulation can also occur unintentionally as a result of overfitting to a given metric~\citep{black2024d,cooper2021hyperparameter,ganesh2024empirical}, underscoring the need for vigilance against the misuse of multiplicity and calling for a more robust operationalization of regulatory frameworks~\citep{black2024d}.

%%%%%%%%%%%%%%%%%%%%%%%%%%%%%%%%%%%%%%
%%%%%%%%%%%%%%%%%%%%%%%%%%%%%%%%%%%%%%

%%%%%%%%%%%%%%%%%%%%%%%%%%%%%%%%%%%%%%
%%%%%%%%%%%%%%%%%%%%%%%%%%%%%%%%%%%%%%

\section{Multiplicity and Responsible AI}
\label{sec:responsibleai}

We now place multiplicity in the broader landscape of responsible AI. In this section, we examine the two key concerns for individuals originating from multiplicity, i.e., arbitrariness in model selection and outcome homogenization.
Again, our discussion here builds on what was termed as the \textit{individual}-level effects of multiplicity by \citet{black2022model}, while also focusing on topics that have gained more interest recently, such as homogenization.

\subsection{Arbitrariness as a Responsible AI Concern}
\label{sec:arbitrariness_concern}

Arbitrary decisions in an automated system can be deeply concerning when they can have direct and lasting impacts on human lives~\citep{black2022model,gomez2024algorithmic,sokol2024cross,watson2024predictive}. Borrowing an analogy from \citet{gomez2024algorithmic}, imagine a judge deciding legal cases by flipping a coin. While this may seem extreme, it demonstrates how models can have arbitrariness embedded in them due to an analogous coin flip done by the developer during model design. This aligns closely with our ICA framework (\S \ref{sec:design_choices}), where we discuss the \textit{arbitrary} choices that can contribute to multiplicity.

There are contexts where a degree of ``controlled randomness'' may be acceptable—or even necessary (\S \ref{sec:homogenization}). However, arbitrariness is a significant concern in scenarios where individuals lack access to other \textit{`equivalent opportunities'}~\citep{barocas-hardt-narayanan}.  For instance, in hiring, some level of arbitrariness may be acceptable, or even necessary to deal with the concerns of homogenization (\S \ref{sec:homogenization}). This is because individuals looking for a job often seek multiple opportunities across companies, increasing their chances of being hired elsewhere. In contrast, domains such as law or medicine typically involve singular, high-stakes decisions with no equivalent alternatives. In such situations, the presence of arbitrariness raises serious concerns for the responsible deployment of machine learning models.

The negative effects of multiplicity extend beyond the arbitrariness of just the final prediction (\S \ref{sec:impact}). For instance, multiplicity in counterfactual explanations can impact the validity of algorithmic recourse~\citep{hasan2022mitigating,leventi2022rashomon,baniecki2024grammar,hamman2023robust,leofante2023counterfactual,jiang2024recourse}. The feasibility or nature of recourse might hinge on \textit{arbitrary} design choices made during model development. These decisions can have real-world implications; for instance, a recourse provided by one model may become invalid if the model is updated, invalidating previous efforts. This inconsistency raises ethical and legal concerns~\citep{kobylinska2024exploration,de2024time}.

\begin{table*}[t!]
    \centering
    % \footnotesize
    \small
    \begin{tabular}{p{4.3cm}p{12.4cm}}
        \toprule
        \textbf{Term} & \textbf{Definition} \\
        \midrule
        Algorithmic Blackballing \citep{ajunwa2020auditing} & \textit{`A worker’s lack of control over the portability of applicant data captured by automated hiring systems […] raising the specter of an algorithmically permanently excluded class of job applicants'} \\[0.2em]
        Algorithmic Monoculture \citep{kleinberg2021algorithmic} & \textit{`The notion that choices and preferences will become homogeneous in the face of algorithmic curation'} \\[0.2em]
        Algorithmic Leviathan \citep{creel2022algorithmic} & \textit{`Automated decision-making systems that make uniform judgments across broad swathes of a sector.'} \\[0.2em]
        Outcome Homogenization \cite{bommasani2022picking} & \textit{`The phenomenon of individuals (or groups) exclusively receiving negative outcomes from all ML models they interact with'} \\[0.2em]
        Generative Monoculture \citep{wu2024generative} & \textit{`A distribution shift [towards less varied outputs] from source data (i.e., human-generated training data) to model-generated data (i.e., model outputs) for a specific task.'} \\[0.2em]
        Algorithmic Pluralism \citep{jain2024algorithmic} & \textit{`A state of affairs in which the algorithms used for decision-making are not so pervasive and/or strict as to constitute a severe bottleneck on opportunity.'} \\
        \bottomrule
    \end{tabular}
    \caption{Various terms used in the homogenization literature.}
    \label{tab:homogenization}
    % \vspace{-2.4em}
\end{table*}

This arbitrariness becomes even more problematic when it disproportionately impacts different individuals, particularly harming underrepresented demographics. As hinted earlier, a significant source of arbitrariness is the model’s lack of ability to learn the underlying distribution. Underrepresented groups often face these information gaps, which can manifest as data scarcity due to limited historical records or a lack of understanding of cultural context for how data relates to predictions~\citep{barocas-hardt-narayanan,hooker2021moving,buolamwini2018gender,birhane2022automating,geirhos2020shortcut}. Research has consistently shown that such disparities exacerbate existing inequalities—whether through arbitrariness, uncertainty, or multiplicity~\citep{farnadiposition}. The resulting disproportionate harms across groups highlight the pressing need to address arbitrariness in critical domains~\citep{gomez2024algorithmic,cooper2024arbitrariness,ali2021accounting}.

\textbf{Responsible AI Constraints and Multiplicity:}
We saw that multiplicity is a critical consideration in responsible AI. Interestingly, several works have also explored how multiplicity interacts with other pillars of responsible AI. Studies have shown that imposing fairness constraints can inadvertently increase multiplicity~\citep{long2024individual,cavus2024experimental}. However, \citet{long2024individual} argue that multiplicity stands outside the fairness-utility trade-off, meaning improvements in fairness do not have to entail increased multiplicity. They demonstrate that multiplicity can often be reduced through techniques like ensembling, while maintaining fairness. Similarly, \citet{kulynych2023arbitrary} explore the interaction between differential privacy and multiplicity, revealing that introducing privacy constraints tends to increase multiplicity. These findings underscore the complex interplay between multiplicity and other responsible AI principles, highlighting the need for further research to understand and navigate these trade-offs effectively.

\subsection{Multiplicity and Homogenization}
\label{sec:homogenization}

As discussed above, the extent to which arbitrariness is problematic often depends on the context, and in certain cases, it might not be a standalone concern. For instance, \citet{creel2022algorithmic} argue that arbitrariness in hiring decisions, by itself, is neither a legal nor moral issue. Instead, they suggest that the absence of arbitrariness across systems could lead to a different concern, creating an `algorithmic leviathan', i.e., the standardization of a single outcome across an entire sector. \citet{kleinberg2021algorithmic} discuss a similar concern in the form of `algorithmic monoculture', which would be particularly problematic in interconnected systems, for instance, when multiple banks assess an individual's creditworthiness, algorithmic monoculture would imply that an individual rejected from one bank would be rejected from all banks. Please refer to Table \ref{tab:homogenization} for an overview of other common terms used in this literature.
% Failures in these ecosystems are magnified because errors from one system can cascade across all others, creating systemic risks.

This phenomenon, known as outcome homogenization, refers to the convergence of decisions due to common design choices across multiple models. In our ICA framework (\S \ref{sec:design_choices}), we had termed these as \textit{conventional} choices.  \citet{bommasani2022picking} shows that outcome homogenization can occur even when different algorithms share only certain components, i.e., homogenization can occur even when only some design choices are \textit{conventional}.

Introducing controlled randomness can mitigate these risks by preventing monocultures. In contexts where arbitrariness in individual decisions is less concerning than homogenization, controlled multiplicity is desirable~\citep{barocas-hardt-narayanan,perry2015may,jain2024scarce}. In such situations, when an \textit{intentional} design choice is not possible, the developers should prefer \textit{arbitrary} choices over \textit{conventional} ones wherever feasible (\S \ref{sec:design_choices}). Despite this idea being widely recognized in the academic literature~\citep{barocas-hardt-narayanan,perry2015may,jain2024scarce}, public perception of intentional randomness in decision-making remains skeptical. A recent study by \citet{meyer2024perceptions} indicates a strong aversion of the end users towards any form of randomization or intentional arbitrariness in automated decision-making. Therefore, fostering greater public awareness about the nuanced impacts of multiplicity is crucial before we can develop and employ potential solutions~\citep{barocas-hardt-narayanan}.

%%%%%%%%%%%%%%%%%%%%%%%%%%%%%%%%%%%%%%
%%%%%%%%%%%%%%%%%%%%%%%%%%%%%%%%%%%%%%

%%%%%%%%%%%%%%%%%%%%%%%%%%%%%%%%%%%%%%
%%%%%%%%%%%%%%%%%%%%%%%%%%%%%%%%%%%%%%

\section{Open Questions and Emerging Trends}

Our systematic survey gives us a unique vantage point to identify and discuss several emerging trends in the field.

\begin{itemize}[leftmargin=*]
    \item \textit{Expanding the scope of multiplicity beyond predictions and explanations.} Our key motivation for broadening the definition of multiplicity (\S \ref{sec:formalizing}) was to incorporate multiplicity beyond predictive and explanatory contexts. Although interest in these aspects is growing, further work is needed to bring them to the forefront. We hope our work will encourage future research that fosters a more holistic understanding of multiplicity's broader impact.
    \item \textit{Cost-effective enumeration of Rashomon sets.} A major challenge in auditing multiplicity lies in the resource-intensive nature of training multiple models to enumerate the Rashomon set. While we discussed several works that improve the efficiency of this enumeration (\S \ref{sec:evaluating_metrics}), the need for further research in this direction remains pressing.
    \item \textit{Mathematical foundations of multiplicity.} Establishing stronger mathematical foundations of multiplicity, for instance, our focus on formalizing the distinction between multiplicity, uncertainty, and bias-variance decomposition (\S \ref{sec:uncertainty}), is essential. Fundamental work on the Rashomon effect is less represented (only $12.5\%$; see Figure \ref{fig:systematic_statistics}), highlighting the opportunities for future work on frameworks that rigorously define and explore multiplicity.
    \item \textit{Multiplicity and its interaction with responsible AI.} Given the conversation of arbitrariness as a concern of responsible model development (\S \ref{sec:arbitrariness_concern}), its interaction with other pillars of responsible AI is warranted. Future research on frameworks that address multiplicity within the broader landscape of responsible AI deployment is needed.
    \item \textit{Interdisciplinary perspective on multiplicity.} We found many studies in this field do not engage with the concerns of multiplicity in real-world settings. 
    % As discussed, arbitrariness can be both a force of good (\S \ref{sec:homogenization}) or a cause of harm (\S \ref{sec:arbitrariness_concern}), depending on the context. 
    Our systematic review shows that only $47.5\%$ of works explicitly engage with responsible AI concerns (see Figure \ref{fig:systematic_statistics}), which we believe to be low given the field's relevance to the responsible deployment of machine learning models. Future collaborative efforts across disciplines are thus crucial.
    \item \textit{Multiplicity and LLMs.} As models continue to scale, new challenges emerge. For instance, we see increasing attention given to the concerns of monoculture and homogenization (\S \ref{sec:homogenization}). Even the evaluation of multiplicity becomes increasingly complex, as training multiple models is often infeasible at this scale. Additionally, we see new dimensions of multiplicity like prompt multiplicity~\citep{ganesh2025rethinking}, preference multiplicity, etc., that require deeper examination.
\end{itemize}

\section{Conclusion}

In this work, we systemized existing knowledge on multiplicity, uncovering interesting trends. One limitation of our study is the evolving terminology within the field--terms such as ``multiplicity’’ and ``Rashomon sets’’ have gained prominence only recently. As a result, our survey may have missed relevant works that did not explicitly use this terminology. Despite this, our efforts to formalize key discussions--the language around developer choices, definitions, and the distinction between multiplicity, uncertainty, and variance--represent a crucial step toward unifying the field. We also explored broad trends related to the real-world impacts of multiplicity, building from existing literature and highlighting overarching themes that extend beyond their originally studied contexts. We hope our work provides a platform that is both accessible to newcomers and valuable to experts, fostering further research in multiplicity.

%%%%%%%%%%%%%%%%%%%%%%%%%%%%%%%%%%%%%%
%%%%%%%%%%%%%%%%%%%%%%%%%%%%%%%%%%%%%%

\section*{Acknowledgements}

We would like to thank Shomik Jain for his feedback on an earlier version of the paper.
Funding support for project activities has been partially provided by Canada CIFAR AI Chair, Google award, CIFAR Catalyst Grant award, FRQNT and NSERC Discovery Grants program. We also express our gratitude to Compute Canada for their support in providing facilities for our evaluations.

\bibliography{aaai25}

%%%%%%%%%%%%%%%%%%%%%%%%%%%%%%%%%%%%%%
%%%%%%%%%%%%%%%%%%%%%%%%%%%%%%%%%%%%%%

\appendix

\section{Systematic Literature Review}
\label{sec:app_systematic}

\begin{table*}[t!]
    \centering
    % \footnotesize
    % \small
    \begin{tabular}{lrrr}
        \toprule
        \textbf{Search Term} & \multicolumn{3}{c}{\textbf{Number of Papers}}\\
         & \textbf{DBLP} & \textbf{ACM Digital Library} & \textbf{Combined (Duplicates Removed)} \\
        \midrule
        \textit{`rashomon'} & 34 & 156 & 178 \\
        \textit{`model multiplicity'} & 66 & 95 & 153 \\
        \textit{`set of good models'} & 11 & 28 & 38 \\[0.4em]
        \multicolumn{4}{r}{Total Number of Papers (Duplicates Removed): \quad\quad\quad 339} \\[0.2em]
        \multicolumn{4}{r}{\textbf{Total Number of Papers (After Manual Filtering): \quad\quad\quad\; 80}} \\
        \bottomrule
    \end{tabular}
    \caption{Paper collection statistics.}
    \label{tab:systematicreview}
\end{table*}

\begin{table*}[t!]
    \centering
    % \footnotesize
    % \small
    \begin{tabular}{p{7.5cm}p{0.5cm}p{7.5cm}}
        \toprule
        \textbf{Inclusion Criteria} & & \textbf{Exclusion Criteria} \\
        \midrule
        - Papers whose central contribution was deeply intertwined with the Rashomon effect or multiplicity. & & - Papers that mentioned the Rashomon effect, but did not engage with it in their main contributions. \\[0.4em]
        - Position papers and surveys that devoted a considerable amount of space to the Rashomon effect or multiplicity (We used one complete section worth of space as our threshold). & & - Papers exploring the Rashomon effect or model multiplicity in domains outside machine learning. The term `model' in these cases did not refer to machine learning models. \\[0.4em]
         & & - Commentaries, extended abstracts, tutorials, complete conference proceedings, and other indexed documents that are not considered research papers. \\[0.4em]
         & & - Duplicates for papers which exist in various repositories under different names, and hence could not be filtered automatically. \\
        \bottomrule
    \end{tabular}
    \caption{Inclusion and exclusion criteria for manual filtering.}
    \label{tab:manualfiltering}
\end{table*}

We conducted a systematic literature review to collect and analyze works related to multiplicity. Our process involved the following steps:

\begin{enumerate}[leftmargin=*]
    \item \textbf{Paper Collection:} We indexed various online repositories using a set of predefined search terms. Details of the search are provided in \S \ref{sec:app_paper_collection}. After filtering for duplicates across various search terms and repositories, we retained a total of 339 papers.
    \item \textbf{Manual Filtering:} We manually reviewed the remaining papers to exclude those not relevant to our study. The exact inclusion criteria are noted in \S \ref{sec:app_manual_filtering}. This manual filtering reduced the dataset to 80 papers.
    \item \textbf{Manual Tagging:} Finally, each paper was assigned tags to identify and emphasize the specific aspects of multiplicity it addressed. The set of tags used and categorization rules are discussed in \S \ref{sec:app_tagging}.
\end{enumerate}

After filtering and tagging the papers, we analyzed various statistics, as shown in the main text. The following sections provide a detailed account of each step of the systematic review process.

\subsection{Paper Collection}
\label{sec:app_paper_collection}

We indexed and collected papers using the search terms `\textit{rashomon}', `\textit{model multiplicity}', and `\textit{set of good models}' from two primary sources: DBLP\footnote{\url{https://dblp.org/}} and the ACM Digital Library\footnote{\url{https://dl.acm.org/}}. To ensure comprehensive coverage, we used the extended search in `The ACM Guide to Computing Literature' within the ACM Digital Library. The end date for our search was 31 Dec 2024, and no restrictions were put on the starting date. Together, DBLP and the ACM Digital Library provide coverage of major machine learning conferences (e.g., NeurIPS, ICML, ICLR, AAAI, ACL, NAACL), leading venues focused on responsible AI (e.g., FAccT, AIES, EAAMO), and archival repositories such as arXiv. The exact number of papers collected from each source and search query is detailed in Table \ref{tab:systematicreview}.

\subsection{Manual Filtering}
\label{sec:app_manual_filtering}

After removing duplicates, we had 339 papers in our dataset. Each paper was then manually reviewed to determine its relevance to our survey. Our focus was on works that deeply engaged with the Rashomon effect and multiplicity within the context of machine learning. Thus, papers examining the Rashomon effect outside of machine learning, investigating multiplicity in other domains, or briefly mentioning the Rashomon effect without further exploring it, were excluded. The detailed inclusion and exclusion criteria are outlined in Table \ref{tab:manualfiltering}.

\subsection{Manual Tagging}
\label{sec:app_tagging}

Finally, after manual filtering, we were left with 80 papers. Each paper was then tagged based on their contributions in the context of multiplicity in machine learning. The tags were created by the authors after a preliminary review of the papers during the manual filtering step. Strict rules were defined for each tag, as shown in Table \ref{tab:manualtagging}. Every paper was assigned all applicable tags through this manual tagging process. These 80 tagged papers formed the basis of the statistics presented in Fig \ref{fig:systematic_statistics} of the main text.

\begin{table*}[!]
    \centering
    % \footnotesize
    % \small
    \begin{tabular}{lp{12cm}}
        \toprule
        \textbf{Tag} & \textbf{Rule} \\
        \midrule
        Rashomon Effect in ML & Papers that provide fundamental insights into the Rashomon effect in ML, including its causes, forms of manifestation, broader characteristics, etc. \\[0.4em]
        Rashomon Set Exploration & Papers that focus on the problem of enumerating the Rashomon set, including papers which highlight properties of the Rashomon set. This does not include papers that already assume access to the Rashomon set. \\[0.4em]
        Better Models and Ensembles & Papers that use the Rashomon set to find better models, or combine multiple models from the Rashomon set to improve certain objectives. Thus, any work that takes advantage of the flexibility the Rashomon set provides. \\[0.4em]
        Hacking Metrics & Papers that highlight the negatives of underspecification, using the Rashomon effect to hack existing metrics, checks, or regulations. \\[0.4em]
        Responsible AI & Papers that explicitly engage with the harms or benefits of the Rashomon effect in the context of Responsible AI. Papers whose central contribution is related to Responsible AI, but do not engage with that aspect of their impact, were not included here. For example, papers tackling explanation multiplicity strictly from a technical point of view were not given this tag. \\[0.4em]
        Application & Papers whose central contribution is limited to a particular application. \\[0.4em]
        Survey & Position papers and surveys that either provide an overarching discussion on multiplicity or place multiplicity in a broader context. \\[0.4em]
        Predictive Multiplicity & Papers that focus on predictive multiplicity, i.e., when models in the Rashomon set have varying predictions. \\[0.4em]
        Explanation Multiplicity & Papers that focus on explanation multiplicity, i.e., when models in the Rashomon set have varying explanations. \\[0.4em]
        Other Multiplicity & Papers that focus on any other form of multiplicity beyond predictive or explanation multiplicity. This can include fairness multiplicity, OOD robustness multiplicity, model complexity multiplicity, feature interaction multiplicity, etc. \\
        \bottomrule
    \end{tabular}
    \caption{Rules for each tag in the manual tagging step.}
    \label{tab:manualtagging}
\end{table*}

\section{Metric Glossary}
\label{sec:app_glossary}

% \paragraph{Notation}

% \begin{itemize}
%     \item $S_\varepsilon$: $\varepsilon$-Rashomon set
%     \item $h_k$: Functional representation of an ML model
%     \item $D$: Dataset on which multiplicity is measured
%     \item $x_k \in D$: A datapoint in the dataset
%     \item $|D| = n$: Dataset size
% \end{itemize}

% \paragraph{Metric Definitions}

\begin{itemize}
    \item Ambiguity~\cite{marx2020predictive}
    % \[
    % => \frac{1}{n} \sum_{i=1}^{n} \max_{h_1, h_2 \in S_\varepsilon} \mathbbm{1}\left[ h_1(x_i) \ne h_2(x_i) \right]
    % \]
    \item Obscurity~\cite{cavus2024experimental}
    \item Discrepancy~\cite{marx2020predictive}
    % \[
    % => \max_{h_1, h_2 \in S_\varepsilon} \left[ \frac{1}{n} \sum_{i=1}^{n} \mathbbm{1}\left[ h_1(x_i) \ne h_2(x_i) \right] \right]
    % \]
    \item Degree of Underspecification~\cite{teney2022predicting}
    \item Viable Prediction Range~\cite{watson2023predictive}
    \item Rashomon Capacity~\cite{hsu2022rashomon}
    \item Multi-target Ambiguity~\cite{watson2023multi}
    \item Rank List Sensitivity~\cite{oh2022rank}
    \item Std. of Scores~\cite{heljakka2022disentangling,long2024individual}
    \item Representational Multiplicity~\cite{heljakka2022disentangling}
    \item Region Similarity Score~\cite{somepalli2022can}
    \item Self-consistency~\cite{cooper2024arbitrariness}
    \item $\epsilon$-robust to Dataset Multiplicity~\cite{meyer2023dataset}
    \item Unfairness Range~\cite{mata2022computing}
    \item Rashomon Ratio~\cite{semenova2022existence}
    \item Underspecification Score~\cite{madras2019detecting}
    \item Accuracy Under Intervention~\cite{ganesh2024empirical}
    \item Consistency~\cite{watson2022agree}
    \item Model Class Reliance~\cite{fisher2019all}
    \item Attribution Agreement~\cite{krishna2022disagreement,brunet2022implications}
    \item Profile Disparity Index~\cite{kobylinska2024exploration}
    \item Inv. Cost of Neg. Surprise~\cite{pawelczyk2020counterfactual}
    \item Variable Importance Clouds~\cite{dong2019variable}
    \item Coverage \& Interval Width~\cite{marx2023but}
\end{itemize}

% \section{Definitions of Rashomon Set and Multiplicity: Example Case Studies}
% \label{sec:app_reducing}

% We provide some examples of existing debates or future directions of research stemming from our generalized definitions of Rashomon set and multiplicity in \S \ref{sec:formalizing}.

% \paragraph{Rashomon Set on Training vs Validation}

% The most common definition of Rashomon set relies on finding 'good models' based on the training loss of models in the hypothesis class~\citep{marx2020predictive}. However, in practice, this is rarely the case, and instead, developers tend to have access to some form of validation set that is used to record model performance and recognize 'good models'. Hence, whether the performance constraint metrics are defined on the train set or the validation set should be clarified, as it can make a big difference in the final Rashomon set. 

% \paragraph{Absolute Threshold vs Relative Threshold}

% Another common tension between existing definitions of Rashomon set is the choice of 

% \paragraph{Rashomon Set with Multiple Performance Constraints}

% \paragraph{Predictive Multiplicity for Binary Classification}

% \paragraph{Predictive Multiplicity on Training vs Test}

% \paragraph{Fairness Multiplicity}

%%%%%%%%%%%%%%%%%%%%%%%%%%%%%%%%%%%%%%
%%%%%%%%%%%%%%%%%%%%%%%%%%%%%%%%%%%%%%

\end{document}